\documentclass{article}

\usepackage{microtype}
\usepackage{graphicx}
\usepackage{subfigure}
\usepackage{booktabs} 

\usepackage{hyperref}

\usepackage[accepted]{icml2024}

\usepackage{amsmath}
\usepackage{amssymb}
\usepackage{mathtools}
\usepackage{amsthm}

\usepackage[capitalize,noabbrev]{cleveref}

\theoremstyle{plain}

\theoremstyle{definition}

\theoremstyle{remark}

\usepackage[textsize=tiny]{todonotes}

\usepackage{tabularx}

\icmltitlerunning{Advancing Explainable AI Toward Human-Like Intelligence: Forging the Path to Artificial Brain}

\begin{document}

\twocolumn[
\icmltitle{Advancing Explainable AI Toward Human-Like Intelligence: Forging the Path to Artificial Brain}



\icmlsetsymbol{equal}{*}

\begin{icmlauthorlist}
  \icmlauthor{Yongchen Zhou}{LIRA}
  \icmlauthor{Richard Jiang}{LIRA}
\end{icmlauthorlist}

\icmlaffiliation{LIRA}{LIRA Center, Lancaster University, Lancaster, England}

\icmlcorrespondingauthor{Yongchen Zhou}{y.zhou52@lancaster.ac.uk}
\icmlcorrespondingauthor{Richard Jiang}{r.jiang2@lancaster.ac.uk}

\icmlkeywords{Explainable AI, Human-Like Intelligence, Artificial General Intelligence}

\vskip 0.3in
]



\printAffiliationsAndNotice{}  

\begin{abstract}

The intersection of Artificial Intelligence (AI) and neuroscience in Explainable AI (XAI) is pivotal for enhancing transparency and interpretability in complex decision-making processes. This paper explores the evolution of XAI methodologies, ranging from feature-based to human-centric approaches, and delves into their applications in diverse domains, including healthcare and finance. The challenges in achieving explainability in generative models, ensuring responsible AI practices, and addressing ethical implications are discussed. The paper further investigates the potential convergence of XAI with cognitive sciences, the development of emotionally intelligent AI, and the quest for Human-Like Intelligence (HLI) in AI systems. As AI progresses towards Artificial General Intelligence (AGI), considerations of consciousness, ethics, and societal impact become paramount. The ongoing pursuit of deciphering the mysteries of the brain with AI and the quest for HLI represent transformative endeavors, bridging technical advancements with multidisciplinary explorations of human cognition.
\end{abstract}
    
\section{Introduction}
\begin{figure}[t]
  \centering
  \includegraphics[width=0.8\linewidth]{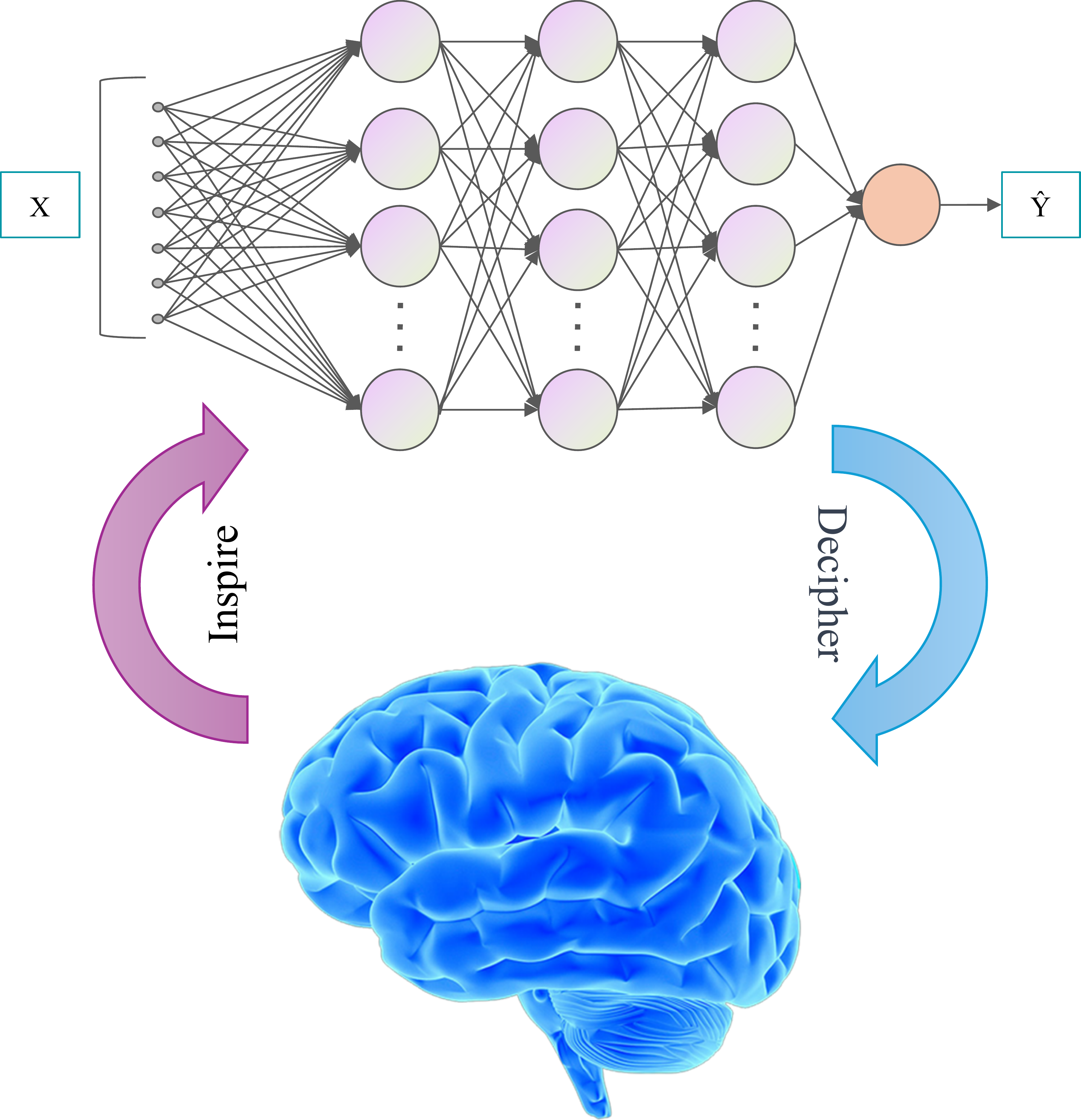}
  \caption{\textbf{The cycle of learning.} As neural networks evolve by mimicking the brain, they offer insights that, in turn, illuminate our understanding of cerebral processes.}
  \label{fig:cycle-of-learning}
\end{figure}

In the rapidly advancing field of Artificial Intelligence (AI), transparency and explainability are crucial, especially in impactful sectors like finance and healthcare \cite{ehsan2021expanding}. The elusive nature of AI's decision-making, often likened to a "black box," emphasizes the critical necessity for Explainable AI (XAI) \cite{saeed2023explainable}. Understanding XAI requires recognizing the iterative nature of AI learning, similar to human brain processes. The visual representation in \cref{fig:cycle-of-learning} captures this dynamic, depicting the reciprocal exchange of insights between neural network development and our understanding of cerebral mechanisms, benefiting both domains.

Contemporary discourse on XAI prompts a reflective analysis of its progression and explainability in current research \cite{ali2023explainable}. Spanning philosophy to computer science, the interdisciplinary nature of this field highlights its complexity \cite{schmid2022missing}. Bengio emphasizes the need for a holistic approach to dispel persistent misconceptions in deep learning \cite{bengio2021deep}. The anthropomorphization of AI introduces ethical concerns, potentially shifting accountability from creators to AI entities \cite{hindennach2023mindful}.

The convergence of XAI with cognitive sciences, especially in pedagogical settings, holds promise for embedding human-like reasoning in AI systems \cite{suffian2023toward}. This interdisciplinary synthesis, including cognitive science and HCI, is crucial for user-centered XAI system design \cite{saeed2023explainable}. These advancements carry transformative implications for sectors like healthcare and the legal system \cite{suffian2023toward}.

This paper delves into the cutting edge of XAI, exploring practical implementations and theoretical foundations. By amalgamating diverse research trajectories, our goal is to cultivate AI systems that replicate human cognitive mechanisms, aligning with the vision of autonomous intelligence \cite{hyder2019artificial}.

\section{The State-of-The-Art XAI}
This section explores contemporary XAI methodologies, emphasizing a spectrum that balances accuracy and interpretability. Refer to \Cref{performance-comparison} for a concise comparative analysis of some methods using the Caltech-101 dataset.

\begin{table*}[t]
  \centering
  \begin{tabular}{l@{\hspace{5pt}}*{3}{@{\hspace{35pt}}c}}
    \toprule
    \textbf{Method} & \textbf{Accuracy} & \textbf{\#Parameters} & \textbf{Interpretability} \\
    \midrule
    SpinalNet \cite{kabir2022spinalnet} & \textbf{97.32\%} & 132,600,000 & \textbf{High} \\
    xDNN \cite{angelov2020towards} & 94.31\% & \textbf{4 per class} & \textbf{High} \\
    VGG-16 \cite{simonyan2014very} & 90.32\% & 138,000,000 & Very low \\
    ResNet-50 \cite{he2016deep} & 90.39\% & 23,000,000 & Very low \\
    Random forest \cite{breiman2001random} & 87.12\% & $\sim$20,000 & Medium \\
    SVM \cite{hearst1998support} & 86.64\% & $\sim$15,000 & Low \\
    kNN \cite{peterson2009k} & 85.65\% & $\sim$300 and all data & Low \\
    Decision tree \cite{quinlan1996bagging} & 86.42\% & $\sim$5 rules per class & \textbf{High} \\
    Naive Bayes \cite{rish2001empirical} & 54.84\% & 409,700 & Medium \\
    \bottomrule
  \end{tabular}
  \caption{\textbf{Performance comparison} for the Caltech-101 dataset. We can get a highly accurate XAI solution that rivals the accuracy achieved by DL.}
  \label{performance-comparison}
\end{table*}

\subsection{Features-Oriented Methods}
Feature-based interpretability techniques, such as Shapley Additive Explanation (SHAP) \cite{lundberg2017unified}, Class Activation Maps (CAMs) \cite{zhou2016learning}, Grad-CAM \cite{selvaraju2017grad}, Grad-CAM++ \cite{chattopadhay2018grad}, Global Attribution Mappings (GAMs) \cite{ibrahim2019global}, and Gradient-based Saliency Maps \cite{simonyan2013deep}, offer diverse insights into machine learning model decision-making.

SHAP employs a game-theoretic approach for both local and global consistency in interpreting feature contributions \cite{lundberg2017unified}. CAMs, tailored for CNNs, use heatmaps to highlight influential image regions \cite{zhou2016learning}, with Grad-CAM and Grad-CAM++ refining this method for increased flexibility and detail \cite{selvaraju2017grad, chattopadhay2018grad}.

GAMs provide a unique perspective by explaining neural network predictions across subpopulations, using a clustering approach to reveal global attribution patterns \cite{ibrahim2019global}. Gradient-based Saliency Maps visualize influential features in image classification by rendering the gradient's absolute value as a heatmap \cite{simonyan2013deep}.

These methods improve understanding of decision locations in inputs but face challenges in explaining 'how' and 'why' decisions are made, particularly in non-additive models and instances with identical objects. Each approach has distinct strengths and limitations, underlining the ongoing effort to enhance accessibility and interpretability in AI decision-making through continuous innovation in feature-based interpretability research.

\subsection{Pixel-Based Methods}
Layer-Wise Relevance Propagation (LRP) \cite{bach2015pixel} illuminates the decision-making process in multilayer neural networks through specific propagation rules, generating a heatmap that visually highlights the contribution of individual pixels to the network's output. It provides insights into the elements positively influencing model decisions and is applicable to pre-trained networks for retrospective feature significance analysis, limited to models compatible with backpropagation. In contrast, DeconvNet \cite{noh2015learning} employs a semantic segmentation approach, using a learned deconvolution network to reveal pixel-level contributions during classification, aligning with transparency efforts. The introduction of a deep belief network \cite{hinton2006fast} further advances interpretability in conventional neural networks. While these methodologies play a pivotal role in demystifying deep learning operations, their effectiveness depends on underlying network structures and the chosen interpretability framework.


\subsection{Concept Models}

AI interpretability has advanced through techniques like Concept Relevance Propagation (CRP) and Concept Activation Vector (CAV) \cite{achtibat2023attribution, kim2018interpretability}. CRP, evolving from Layer-wise Relevance Propagation (LRP) \cite{bach2015pixel}, extends analysis beyond identifying important image sections, offering insights into fundamental concepts shaping AI decisions, especially in image recognition and complex language models like ChatGPT. It provides clarity on the "why" behind AI judgments.

Meanwhile, CAV provides a global perspective on neural networks by associating high-level latent features with human-comprehensible concepts \cite{achtibat2023attribution}. It quantifies the alignment of these features with user-defined concepts, revealing potential flaws in the model's learning process. Automatic concept-based explanations refine this approach by autonomously generating CAVs, reducing human bias. However, the interpretative value depends on the distinctiveness and class-specific relevance of the concepts, posing challenges in application \cite{kim2018interpretability, ghorbani2019towards}.

\subsection{Surrogate Models}
The development of model-agnostic explanation techniques represents a significant milestone in the field of XAI, providing a universal framework to decipher the inner workings of black-box models. Among these, Sparse Linear Subset Explanation (SLISE) \cite{bjorklund2023slisemap} and Local Interpretable Model-Agnostic Explanations (LIME) \cite{dieber2020model} stand out for their unique approaches to demystifying complex machine learning predictions.

SLISE stands out for providing interpretable insights into individual predictions without synthetic samples, enhancing clarity and applicability across diverse machine learning scenarios \cite{bjorklund2023slisemap}. This is particularly valuable in sectors requiring transparency, emphasizing SLISE's pivotal role in advancing interpretable AI.

LIME broadens model-agnostic explanations by focusing on localized, optimized interpretations. It trains an interpretable surrogate model to approximate the local decision-making process, dissecting input images into superpixels to analyze feature impact. While effective, LIME's explanations depend on judicious perturbation parameter selection, highlighting the nuanced balance between heuristic-driven customization and meaningful, human-comprehensible explanations \cite{dieber2020model}.

\subsection{Human-Centric Methods}
Current XAI methods, while providing valuable insights into machine learning models, often fall short in delivering inherently understandable explanations to humans. These techniques typically offer a superficial glimpse into the "black box" of AI, focusing on post hoc explanations related to feature importance or image localities \cite{saeed2023explainable}. This approach diverges significantly from human cognitive processes, including reasoning, making associations, evaluating similarities, and drawing analogies, crucial for various professional domains such as medicine, finance, and law. Essentially, traditional XAI methods have not adequately addressed the underlying structural and parametric intricacies of models in relation to problem-solving, overlooking the critical element of human-like reasoning \cite{hassabis2017neuroscience}.

In contrast, a fundamentally different perspective on explainability has been introduced, viewing it through a human-centric lens. This approach emphasizes understanding and comparing complex entities (such as images, songs, and movies) as whole units rather than breaking them down into isolated features or pixels. It advocates for a methodology where humans relate new information to previously encountered and cognitively processed prototypes, a concept supported by both theoretical and empirical research \cite{angelov2020towards, bien2011prototype}. Unlike statistical methods that rely on averages, this human-centric view aligns with how individuals naturally categorize and comprehend the world around them \cite{bishop2006pattern}.

Recent developments in XAI underscore the importance of tailoring evaluation strategies to how AI systems enhance human-AI interaction, especially in decision-making scenarios. These strategies prioritize the assessment of AI's ability to integrate seamlessly into human workflows, thereby measuring the effectiveness of XAI not just on a technical level but in terms of its accessibility and utility to non-specialists \cite{doshi2017towards}. This shift towards application-specific and human-oriented evaluation frameworks marks a critical step in ensuring that AI technologies serve as comprehensible and collaborative tools in real-world applications, bridging the gap between complex AI algorithms and human interpretability.

\subsection{Usability of Counterfactual Explanations}
Counterfactual explanations are increasingly acknowledged for their ability to elucidate AI decision-making by illustrating how changes in inputs influence outcomes. Frameworks like Alien Zoo \cite{kuhl2023let} are utilized to assess the effectiveness and user-friendliness of these explanations. By providing alternate scenarios, counterfactuals enhance user comprehension of AI systems, making them more approachy accessible and interpretable. Alien Zoo's framework, in particular, is instrumental in assessing how these explanations are received and understood by users, guiding the development of more intuitive XAI systems \cite{wachter2017counterfactual}.

\section{Current Challenges in XAI}
As machine learning models become prevalent in diverse sectors, the need for XAI becomes crucial. XAI aims to make these models transparent, accountable, and understandable to a broad audience, not just technical experts. This shift towards explainability seeks to improve model performance and accessibility, promoting a more democratic approach to AI. However, achieving this goal faces challenges, including technical obstacles and ethical considerations \cite{arrieta2020explainable, gilpin2018explaining}.

\subsection{Explainability of Generative Models}

\begin{figure}[t]
  \centering
  \includegraphics[width=1.0\linewidth]{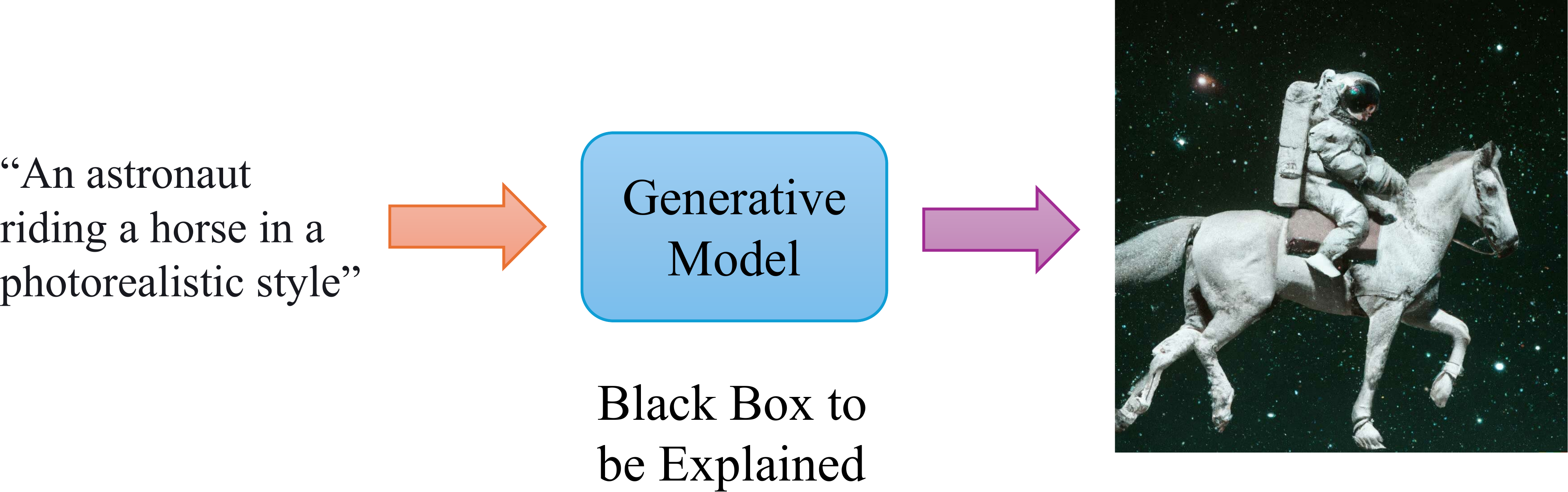}
  \caption{\textbf{The inscrutability of generative models.} Generative models, similar to black boxes, conceal the intricate processes behind their creative outputs, making their internal workings enigmatic and challenging for humans to decipher.}
  \label{fig:generative-models}
\end{figure}

The exploration of generative models in AI, as depicted in \cref{fig:generative-models}, underscores a significant leap forward in data synthesis and creative output generation. However, the complexity and opacity of these models present profound challenges in understanding and explaining their internal decision-making processes.

\textbf{Generative adversarial network.} Generative Adversarial Networks (GANs) have revolutionized data generation with their dual network architecture of a generator and discriminator. Yet, the complexity in interpreting GANs poses significant challenges. Their non-linear, high-dimensional structures, combined with the dynamic adversarial training, make the understanding of their decision-making process complex. Recent studies continue to address these challenges, focusing on enhancing the transparency of GANs \cite{wang2019generative, bau2020understanding}.

\textbf{Neural radiance field.} Neural Radiance Field (NeRF) has marked a significant breakthrough in 3D modeling from 2D images. Understanding the workings of NeRF models is complex due to their processing of high-dimensional data and opaque methods of reconstructing spatial information. The lack of intuitive interpretability in NeRF models adds to the challenge, drawing parallels to the complexity of human brain processes in visual information decoding \cite{mildenhall2021nerf, tewari2022advances, martin2021nerf}.

\textbf{Diffusion model.} Diffusion models, at the forefront of generative AI for image and audio synthesis, encounter substantial explainability challenges. The complexity of their iterative processes, akin to a multi-step chemical reaction, makes understanding these models a daunting task. Recent advancements in diffusion models aim to unravel these complexities \cite{ho2020denoising, dhariwal2021diffusion}.

\textbf{Differential privacy.} The integration of differential privacy in generative models, while ensuring data privacy, also introduces explainability challenges. Balancing transparency with privacy protection is a complex endeavor, often leading to obscured understandings of the models' inner workings \cite{abadi2016deep, jayaraman2019evaluating, li2019differentially}.

\textbf{Large language model.} Large language models (LLMs) such as GPT-4 demonstrate remarkable abilities in text generation and comprehension. However, the complexity of their decision-making processes, involving millions of parameters, presents significant explainability challenges. Understanding the rationale behind specific outputs or language comprehension in LLMs is as intricate as mapping neural pathways in the human brain \cite{brown2020language, bender2021dangers, linzen2020can}.

\subsection{Responsible AI}

Embedding complex human values and ethics in AI systems is a profound challenge, acknowledging the subjective and culturally dependent nature of these concepts. Developing ethically aligned AI requires a deep understanding of diverse cultural and moral frameworks \cite{mittelstadt2019principles, russell2019human}. Responsible AI necessitates transparent systems that can clearly articulate their reasoning to build user trust in ethical decision-making \cite{arrieta2020explainable}.

The scrutiny of fairness in AI aims to detect and neutralize biases to prevent the perpetuation of social inequalities \cite{mehrabi2021survey}. Accountability is crucial, requiring structures to hold AI systems and developers responsible for outcomes, guided by clear ethical guidelines and remediation avenues when deviations from societal norms occur \cite{dignum2019responsible}.

Strengthening the ethical foundation of AI demands interdisciplinary collaboration across technology, humanities, and social sciences. This collaborative approach ensures the construction of AI systems that excel not only in task performance but also align with the ethical and cultural dimensions of the societies they serve.

\subsection{Ethical Implications of Explanations}
The ethical landscape of XAI is complex, emphasizing the importance of addressing moral and societal impacts as AI systems, especially LLMs, advance. Key ethical considerations include the need for unbiased AI operations to ensure fair and just decisions across all demographics \cite{baniecki2021dalex, bellamy2018ai}. This fairness is crucial for establishing an ethical foundation in AI development. Transparency in AI mechanisms is essential for building user trust, especially in sensitive domains where ethical integrity is paramount \cite{slack2021reliable, mehrabi2021survey}. Ongoing discussions in the field highlight the intricate responsibilities associated with technological advancements, emphasizing the requirement for a robust ethical framework as AI becomes deeply integrated into society \cite{rahwan2019machine}. The discourse on the moral and ethical implications of advanced AI systems underscores the need for continuous exploration and dialogue to navigate evolving ethical challenges \cite{bostrom2014superintelligence, russell2019human, wallach2008moral}.
\section{From XAI to Artificial Brain}
\subsection{Understanding and Mimicking Brain Functionality}



\textbf{Challenges in Brain Complexity.} Replicating the intricate neural processes of the human brain remains a significant hurdle for AI systems, as evident from recent studies highlighting the difficulty in simulating dynamic neural interplay \cite{prieto2016neural, friston2005models, hassabis2017neuroscience}.

\textbf{Limitations of Neural Networks.} Despite progress, current neural networks fall short in emulating the depth of learning and adaptability seen in the human brain, posing a hurdle in achieving true intelligence in AI systems \cite{saleem2022comparative, lecun2015deep, marblestone2016toward}.

\textbf{Explicability in Deep Learning.} Deep learning models, often seen as black boxes, present a significant challenge in explicability. Balancing advanced capabilities with the need for understanding their decision-making process is a focal point in Explainable AI (XAI) research, particularly important for applications demanding trust and transparency \cite{samek2019explainable, arrieta2020explainable}.

\subsection{AI Consciousness and Cognition}
\textbf{Theories of AI consciousness.} The exploration of consciousness through frameworks such as Integrated Information Theory (IIT) and Attention Schema Theory (AST) offers profound insights into the essence of awareness. IIT posits that consciousness emerges from the capacity of a system to process integrated information \cite{tononi2016integrated}, whereas AST considers consciousness as a byproduct of the brain's model of attention \cite{graziano2017attention}. These theories provide a foundational understanding that could extend to the realm of artificial intelligence, suggesting pathways to interpret AI behaviors within the context of XAI.

\textbf{AI sentience and cognition.} The ongoing debate surrounding AI's potential for sentience, especially within the ambit of LLMs, enriches the discourse on consciousness. This conversation spans the ethical, philosophical, and practical dimensions of endowing AI with 'sentience,' pondering over the potential rights and duties this might imply \cite{chamola2023review}. As LLMs display increasingly complex behaviors, the line between programmed responses and genuine cognitive processes becomes blurred, presenting a unique challenge for XAI in demarcating clear explanations for AI actions that mimic conscious decisions \cite{dehaene2014consciousness, bender2021dangers}.

\textbf{Theory of mind in AI.} The progression of LLMs towards understanding scenarios indicative of a Theory of Mind (ToM) underscores the sophistication of AI cognitive models. This evolution provokes pivotal inquiries about AI's level of 'understanding' and its implications for developing XAI frameworks that can explain AI decisions in human-centric terms \cite{premack1978does, tsoukalas2018theory}.

\textbf{AI vs human cognition.} The comparison between the cognitive processes of LLMs and humans remains a fertile area of inquiry. Investigating how AI systems and humans differ in processing information and interpreting emotions highlights the complexities involved in making AI's decision-making processes transparent and understandable through XAI \cite{lake2017building, marcus2018deep}.

\textbf{Evaluating AI consciousness.} Crafting metrics to evaluate AI consciousness and its understanding of ToM is an emerging challenge. This endeavor necessitates interdisciplinary collaboration, drawing on neuroscience, psychology, and computer science to forge evaluation tools that not only assess AI's cognitive capabilities but also its ability to make decisions in a manner that is explainable and interpretable to humans \cite{gamez2021measuring}. The quest for such metrics is integral to advancing XAI, ensuring that as AI systems grow more complex and ostensibly 'conscious,' they remain accountable and comprehensible to the people who use them.

\textbf{Evaluating AI Consciousness.} Establishing metrics for assessing AI consciousness and ToM understanding poses an interdisciplinary challenge. Collaboration across neuroscience, psychology, and computer science is essential to create tools that evaluate AI's cognitive capabilities while ensuring accountability and interpretability in XAI as AI systems become more complex and seemingly 'conscious.'

\subsection{Emotional AI Evolution}
\textbf{Emotional Intelligence in AI.} Efforts to integrate emotional intelligence into AI, particularly large language models, mark a significant shift towards recognizing the role of emotions in human cognition. This development aims to bridge the gap between human-machine interaction, promising a more relatable and intuitive user experience. However, ethical considerations arise regarding the authenticity of AI's emotional understanding and potential misperceptions about its empathetic capabilities \cite{mcduff2018designing, rahwan2019machine}.

\textbf{Algorithmic Emotional Complexity.} As AI evolves to simulate human emotions, algorithms must increasingly consider cultural and situational nuances. Recognizing and responding appropriately to the variability and subjectivity of human emotions requires advanced programming approaches. XAI plays a crucial role in making emotional AI's underlying mechanisms transparent and understandable to users, addressing the need for a sophisticated framework \cite{zhan2023multimodal, mcstay2018emotional}.

\textbf{Simulated Empathy Limitations.} AI's attempts to simulate empathy highlight the fundamental absence of genuine consciousness. Recognizing this distinction is crucial for setting realistic expectations and ensuring ethically responsible deployment of emotional AI. XAI contributes by clarifying the extent and limitations of AI's emotional intelligence, providing insights into how AI interprets and reacts to emotional cues \cite{mcduff2018designing}.

\textbf{Emotional AI in Healthcare.} The use of emotional AI in healthcare offers personalized emotional support, enhancing patient care. However, challenges arise from depersonalized interactions and the risk of oversimplifying human emotions, impacting patient trust. XAI becomes essential in elucidating the decision-making processes of emotional AI systems, aligning technology more closely with patient needs and expectations in healthcare settings \cite{riek2017healthcare}.

\subsection{The Personality of AI}

\textbf{Conceptualizing AI Personality.} Exploring AI's embodiment of human-like traits raises interdisciplinary questions in psychology, AI ethics, and human-computer interaction \cite{gerrish2018smart, goleman2020emotional}. As AI integrates into social and professional realms, the concept of AI personalities influences user experience, prompting considerations about transparency and predictability. XAI emerges as a crucial framework for understanding how personality traits impact AI decision-making.

\textbf{Technical Aspects of AI Personality Simulation.} Examining algorithms and methodologies central to AI's interactive capabilities \cite{mcduff2018designing, segal2018gender} aims to refine AI personalities for enhanced relatability and engagement. From an XAI perspective, elucidating technical mechanisms is vital to demystify how personality traits are modeled and manifested in AI interactions, ensuring not only relatability but also comprehensibility of AI responses and decisions.

\textbf{Practical Implications and Case Studies.} Real-world examples and case studies shed light on the practical implications of infusing personality into AI systems \cite{luxton2014artificial, wellman2014making}. Analyzing these cases through an XAI lens provides insights into how personalities impact interpretability and user experience. Understanding these outcomes helps craft AI personalities that enhance transparency and rationale, aligning with XAI's goal of making AI decisions more interpretable and justifiable to users.

\subsection{Creating Biologically Plausible AI Models}


\textbf{Bridging the Biological-Computational Model Gap.} Closing the disparity between biological neural networks and computational AI models is a formidable challenge for advancing artificial intelligence. Significant differences in information processing highlight the complexity of replicating biological functions in AI systems \cite{gherman2023bridging, hassabis2017neuroscience, marblestone2016toward}, offering insights for improving AI interpretability and transparency, crucial aspects of XAI.

\textbf{Complexity in Biological Integration.} Developing AI models mirroring biological neural networks is intricate. These networks balance electrical and chemical signals, adapt through learning, and dynamically reconfigure—processes challenging to emulate in computational algorithms. Achieving biologically plausible models demands a multidisciplinary approach, combining insights from neuroscience, cognitive science, and computer engineering to closely mirror the nuanced functionality of the human brain \cite{hassabis2017neuroscience, marblestone2016toward, dean20201}.

\subsection{Human-AI Interaction and Cognitive Alignment}

\textbf{AI Communication.} Bridging the gap between AI processing and human communication styles is a key challenge \cite{garcia2019human, clark2019makes}.

\textbf{Human-Centric XAI.} Successful XAI relies on a human-centric design, prioritizing intuitive interfaces and easily understandable explanations for broader accessibility \cite{baniecki2021dalex, miller2019explanation, shneiderman2022human}.

\textbf{Learning from Human Cognition.} Real-world XAI applications demand not only technically accurate but also comprehensible explanations, particularly in decision-making contexts \cite{wang2022advanced, jacovi2020towards, wiegreffe2019attention, pope2019explainability, yuan2022explainability}.

\textbf{Precision-Interpretability Balance.} Achieving a balance between precision and interpretability is crucial in XAI, especially in fields like healthcare and law, where transparent decision-making is paramount \cite{doshi2017towards, gunning2019xai, fiorucci2020machine}.

\subsection{Learning from the Brain to Enhance AI}
\textbf{Neuroscientific principles in XAI.} As the demand for transparency in AI systems grows, so does the field of XAI \cite{adadi2018peeking}. Techniques like Local Interpretable Model-Agnostic Explanations (LIME) \cite{ribeiro2016should} and Shapley Additive Explanations (SHAP) \cite{lundberg2017unified} have been instrumental in demystifying the decision-making processes of complex models. Additionally, methods such as Layer-wise Relevance Propagation (LRP) \cite{bach2015pixel} and Integrated Gradients (IG) \cite{sundararajan2017axiomatic} offer deeper insights into the influence of input features in specific model architectures. These methods can be seen as parallel to how neuroscientists attempt to decipher neural network activities in the brain.

\textbf{Incorporating neuroscientific insights.} Integrating findings from neuroscience into AI development is challenging due to the complex nature of neural mechanisms, as highlighted in recent interdisciplinary studies \cite{marcus2018deep, hassabis2017neuroscience}.

\textbf{Bridging research fields.} The challenge of bridging methodologies and terminologies between AI and neuroscience remains significant, as reported in recent collaborative efforts \cite{yamins2016using, kriegeskorte2015deep}.

\textbf{Neurodiversity and XAI adaptability.} XAI's adaptability is tested across various domains, each with unique requirements for interpretability. In healthcare, for instance, explainable models can aid clinicians in understanding AI-assisted diagnoses \cite{holzinger2017we}. In finance, they can help in clarifying credit scoring models for both providers and consumers \cite{weerts2023fairlearn}. Additionally, XAI's role is becoming increasingly important in fields like autonomous driving and environmental modeling, where decisions have significant implications for safety and sustainability. Customizing XAI tools to meet these diverse needs is paramount for their effective integration into different sectors.

\textbf{Real-world XAI applicability.} XAI methodologies must prove their worth in real-world scenarios \cite{baniecki2021dalex}. Their applicability and scalability are crucial, especially when dealing with large and complex AI systems \cite{hedstrom2023quantus}. Scalable XAI solutions are essential for widespread adoption, ensuring that interpretability does not come at the cost of reduced performance or increased resource demands.

\textbf{AI visualization and brain imaging.} Visualization techniques such as Saliency Maps and Gradient Input are crucial in enhancing the interpretability of AI models \cite{simonyan2013deep, shrikumar2017learning}, similar approaches are used in brain imaging (e.g., fMRI, PET scans) to understand which areas of the brain are activated during specific tasks, offering a window into the brain's decision-making processes.

\textbf{Cognitive neuroscience in XAI evaluation.} The evaluation of XAI methodologies often involves sensitivity analysis to assess their robustness \cite{baehrens2010explain}. Moreover, incorporating user studies is becoming essential in evaluating the effectiveness of XAI, especially in terms of how different stakeholders perceive and interact with AI explanations \cite{abdul2020cogam}. This analysis helps in determining the reliability of different XAI methods, identifying their strengths and limitations \cite{alvarez2018robustness, adebayo2018local, adebayo2018sanity}.

\section{Futher Discussion}
\subsection{Artificial General Intelligence}
\begin{figure}[t]
  \centering
  \includegraphics[width=0.75\linewidth]{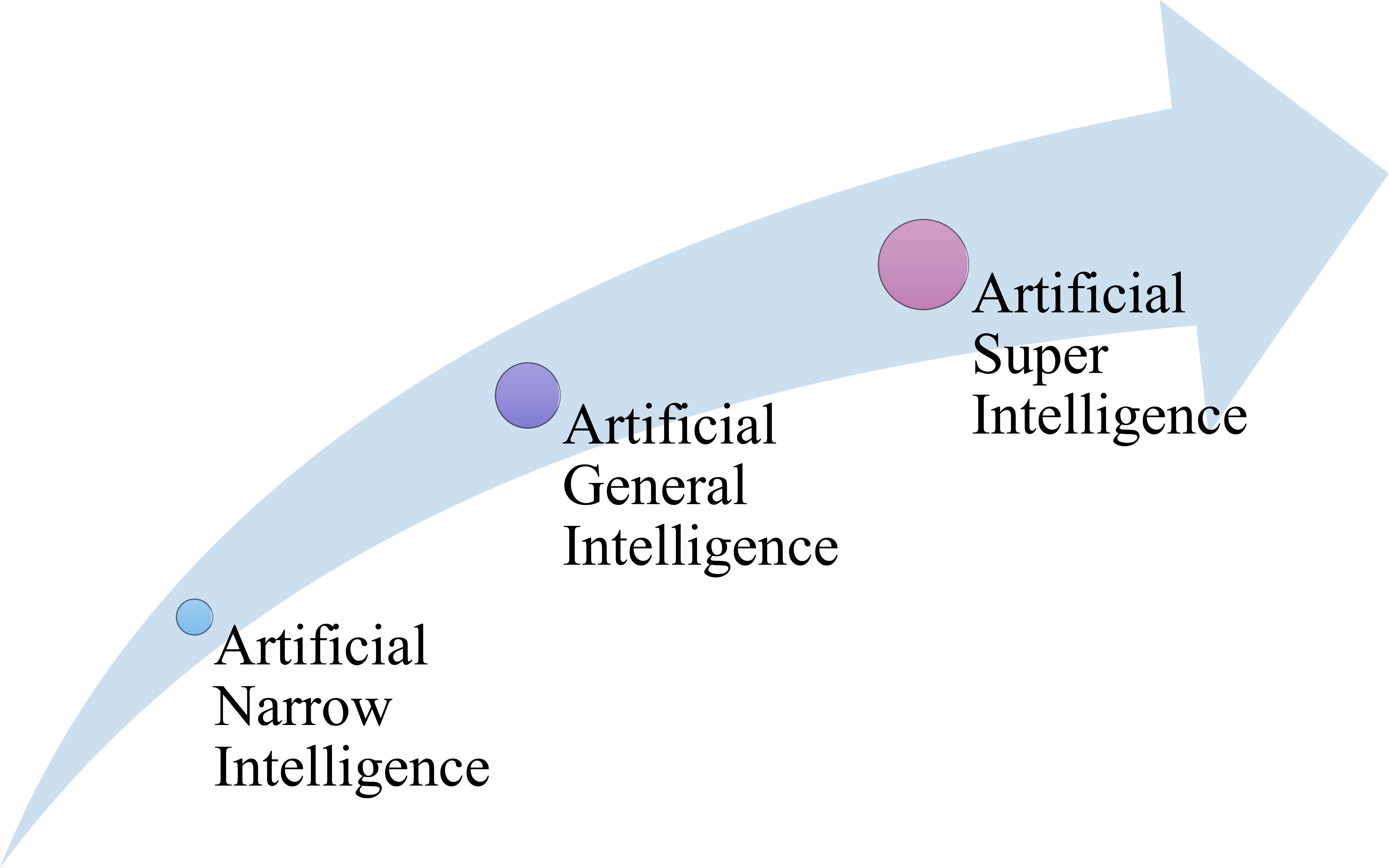}
  \caption{\textbf{The evolution of artificial intelligence.} The arrow's ascent reflects the dynamic growth and expanding capabilities of AI, marking key developments from basic algorithms to advanced sentient systems.}
  \label{fig:evolution}
\end{figure}

Embarking on the journey towards Artificial General Intelligence (AGI) aims to create systems with human-like cognition, surpassing narrow tasks, as shown in \cref{fig:evolution}. This involves enhancing machine learning with advanced neural networks, especially hybrid models blending deep learning and symbolic reasoning \cite{zhang2023toward}.

Integrating emotional intelligence into AGI enhances human-AI interactions through affective computing \cite{zall2022comparative}. Ethical considerations are crucial, requiring comprehensive guidelines and governance frameworks to address AI consciousness, rights, and societal impact \cite{mclean2023risks}.

Scalability and energy efficiency are key research areas for AGI systems. The focus is on developing intelligent, adaptable, scalable, and energy-efficient systems using innovative hardware technologies and computational techniques \cite{moura2023scalable}. Addressing these challenges is essential for advancing AGI ethically and responsibly for societal benefit.

\subsection{Neuro-AI Interface}

The Neuro-AI Interface sector is undergoing a transformative convergence of neuroscience and artificial intelligence, ushering in an era of innovation to decode and translate neural signals, bridging the gap between biology and digital realms. This interdisciplinary junction holds promise for groundbreaking advancements in healthcare, technology, and cognitive science, significantly improving our understanding of cerebral mechanics \cite{wang2020synthetic}. In neuroprosthetics, these interfaces are revolutionizing the restoration and augmentation of motor, sensory, and cognitive functions, granting individuals with disabilities unprecedented interaction capabilities \cite{donoghue2002connecting}. Moreover, they play a crucial role in unraveling brain functionalities and disorders, leveraging AI's analytical capabilities to extract novel insights from intricate neural datasets. This has the potential to revolutionize treatments for neurological disorders and enhance brain imaging technologies \cite{fetz2007volitional, moxon2015brain}. As we approach the ability to control devices and communicate through thoughts, neuro-AI interfaces not only challenge existing interaction paradigms but also pose ethical and philosophical questions about privacy, autonomy, and the essence of human identity in an increasingly interconnected world \cite{ritchie2019decoding}.

\subsection{Deciphering the Brain Mysteries with AI}
The imminent fusion of AI and neuroscience is poised to unlock the profound mysteries of the human brain \cite{hassabis2017neuroscience}. Leveraging AI's capacity to emulate and simulate neural processes, researchers stand on the brink of delving into the brain's intricate mechanisms. This venture goes beyond merely replicating human cognitive processes in AI systems; it aims to decode the brain's own language—interpreting the complex, dynamic patterns of neuronal communication that underpin thought \& behavior.

Central to this interdisciplinary collaboration are several pivotal areas of exploration. For instance, AI's insights into memory formation could dramatically transform our understanding of and approach to learning and memory disorders, shedding light on conditions such as Alzheimer's disease with potential for groundbreaking treatments \cite{hassabis2017neuroscience}. Moreover, AI's simulation of learning paradigms may offer new perspectives on cognitive acquisition, processing, and retention, heralding innovative educational strategies and cognitive enhancement approaches \cite{lake2017building}.




This convergence holds great promise in exploring the origins of consciousness, with AI's analytical prowess positioned as a crucial tool in unraveling the intricate neuronal processes leading to consciousness. This endeavor merges computational neuroscience with philosophical discourse, addressing fundamental existential questions \cite{dehaene2014consciousness}. Additionally, the synergy between AI and neuroscience is driving advancements in neurological healthcare, leveraging AI's simulation of neural networks for more effective interventions, including innovative therapies like brain-computer interfaces and neurostimulation techniques. The integration of computational neuroscience with advanced AI modeling marks a new era, endowing AI systems with human-like cognitive capabilities and accelerating progress in AI while providing transformative insights into neurological health and human intellect augmentation \cite{wang2020synthetic}.

\subsection{Human-Like Intelligence}
Human-Like Intelligence (HLI) in AI research aims to instill machines with cognitive and emotional capabilities resembling those of humans, delving into consciousness, self-awareness, and emotional intelligence \cite{assran2023self}. The exploration of consciousness involves creating systems with self-awareness, enabling informed decision-making while posing technical challenges and raising philosophical and ethical questions \cite{dehaene2014consciousness, tononi2016integrated}.

HLI highlights emotionally intelligent AI, integrating natural language processing, emotion recognition, and context-aware computing \cite{assran2023self}. This has transformative potential in areas such as customer service, mental health, and personal assistance, fostering natural, empathetic interactions. The synergy of consciousness and emotional intelligence in AI signifies a significant stride in replicating human intricacies. HLI's development goes beyond technicalities, involving a multidisciplinary exploration of human thought processes, emotional responses, and consciousness. This convergence of AI, neuroscience, psychology, and ethics seeks to create AI systems that not only emulate but also comprehend and relate to the human experience fully \cite{dehaene2014consciousness}.
\section{Conclusion}


In conclusion, the paper underscores the pivotal role of Explainable AI (XAI) in elucidating AI decision-making processes, addressing challenges in generative models, responsible AI practices, and ethical considerations. As Artificial General Intelligence (AGI) becomes a focus, consciousness, ethics, and societal impacts gain prominence. The convergence of AI and neuroscience not only contributes to neurological healthcare but also propels the development of AI systems mirroring human cognition. Additionally, the exploration of Human-Like Intelligence (HLI) in AI opens new possibilities in emotional intelligence and human-like interactions, shaping the future trajectory of AI towards transparency, interpretability, and more nuanced understanding of the human experience. The ongoing multidisciplinary efforts in deciphering brain mysteries with AI highlight the transformative potential of bridging technical advancements with insights from neuroscience, psychology, and ethics to create AI systems that emulate and comprehend the intricacies of human intelligence.

\bibliography{main}

\begin{thebibliography}{123}
\providecommand{\natexlab}[1]{#1}
\providecommand{\url}[1]{\texttt{#1}}
\expandafter\ifx\csname urlstyle\endcsname\relax
  \providecommand{\doi}[1]{doi: #1}\else
  \providecommand{\doi}{doi: \begingroup \urlstyle{rm}\Url}\fi

\bibitem[Abadi et~al.(2016)Abadi, Chu, Goodfellow, McMahan, Mironov, Talwar, and Zhang]{abadi2016deep}
Abadi, M., Chu, A., Goodfellow, I., McMahan, H.~B., Mironov, I., Talwar, K., and Zhang, L.
\newblock Deep learning with differential privacy.
\newblock In \emph{Proceedings of the 2016 ACM SIGSAC conference on computer and communications security}, pp.\  308--318, 2016.

\bibitem[Abdul et~al.(2020)Abdul, von~der Weth, Kankanhalli, and Lim]{abdul2020cogam}
Abdul, A., von~der Weth, C., Kankanhalli, M., and Lim, B.~Y.
\newblock Cogam: measuring and moderating cognitive load in machine learning model explanations.
\newblock In \emph{Proceedings of the 2020 CHI Conference on Human Factors in Computing Systems}, pp.\  1--14, 2020.

\bibitem[Achtibat et~al.(2023)Achtibat, Dreyer, Eisenbraun, Bosse, Wiegand, Samek, and Lapuschkin]{achtibat2023attribution}
Achtibat, R., Dreyer, M., Eisenbraun, I., Bosse, S., Wiegand, T., Samek, W., and Lapuschkin, S.
\newblock From attribution maps to human-understandable explanations through concept relevance propagation.
\newblock \emph{Nature Machine Intelligence}, 5\penalty0 (9):\penalty0 1006--1019, 2023.

\bibitem[Adadi \& Berrada(2018)Adadi and Berrada]{adadi2018peeking}
Adadi, A. and Berrada, M.
\newblock Peeking inside the black-box: a survey on explainable artificial intelligence (xai).
\newblock \emph{IEEE access}, 6:\penalty0 52138--52160, 2018.

\bibitem[Adebayo et~al.(2018{\natexlab{a}})Adebayo, Gilmer, Goodfellow, and Kim]{adebayo2018local}
Adebayo, J., Gilmer, J., Goodfellow, I., and Kim, B.
\newblock Local explanation methods for deep neural networks lack sensitivity to parameter values.
\newblock \emph{arXiv preprint arXiv:1810.03307}, 2018{\natexlab{a}}.

\bibitem[Adebayo et~al.(2018{\natexlab{b}})Adebayo, Gilmer, Muelly, Goodfellow, Hardt, and Kim]{adebayo2018sanity}
Adebayo, J., Gilmer, J., Muelly, M., Goodfellow, I., Hardt, M., and Kim, B.
\newblock Sanity checks for saliency maps.
\newblock \emph{Advances in neural information processing systems}, 31, 2018{\natexlab{b}}.

\bibitem[Ali et~al.(2023)Ali, Abuhmed, El-Sappagh, Muhammad, Alonso-Moral, Confalonieri, Guidotti, Del~Ser, D{\'\i}az-Rodr{\'\i}guez, and Herrera]{ali2023explainable}
Ali, S., Abuhmed, T., El-Sappagh, S., Muhammad, K., Alonso-Moral, J.~M., Confalonieri, R., Guidotti, R., Del~Ser, J., D{\'\i}az-Rodr{\'\i}guez, N., and Herrera, F.
\newblock Explainable artificial intelligence (xai): What we know and what is left to attain trustworthy artificial intelligence.
\newblock \emph{Information Fusion}, 99:\penalty0 101805, 2023.

\bibitem[Alvarez-Melis \& Jaakkola(2018)Alvarez-Melis and Jaakkola]{alvarez2018robustness}
Alvarez-Melis, D. and Jaakkola, T.~S.
\newblock On the robustness of interpretability methods.
\newblock \emph{arXiv preprint arXiv:1806.08049}, 2018.

\bibitem[Angelov \& Soares(2020)Angelov and Soares]{angelov2020towards}
Angelov, P. and Soares, E.
\newblock Towards explainable deep neural networks (xdnn).
\newblock \emph{Neural Networks}, 130:\penalty0 185--194, 2020.

\bibitem[Arrieta et~al.(2020)Arrieta, D{\'\i}az-Rodr{\'\i}guez, Del~Ser, Bennetot, Tabik, Barbado, Garc{\'\i}a, Gil-L{\'o}pez, Molina, Benjamins, et~al.]{arrieta2020explainable}
Arrieta, A.~B., D{\'\i}az-Rodr{\'\i}guez, N., Del~Ser, J., Bennetot, A., Tabik, S., Barbado, A., Garc{\'\i}a, S., Gil-L{\'o}pez, S., Molina, D., Benjamins, R., et~al.
\newblock Explainable artificial intelligence (xai): Concepts, taxonomies, opportunities and challenges toward responsible ai.
\newblock \emph{Information fusion}, 58:\penalty0 82--115, 2020.

\bibitem[Assran et~al.(2023)Assran, Duval, Misra, Bojanowski, Vincent, Rabbat, LeCun, and Ballas]{assran2023self}
Assran, M., Duval, Q., Misra, I., Bojanowski, P., Vincent, P., Rabbat, M., LeCun, Y., and Ballas, N.
\newblock Self-supervised learning from images with a joint-embedding predictive architecture.
\newblock In \emph{Proceedings of the IEEE/CVF Conference on Computer Vision and Pattern Recognition}, pp.\  15619--15629, 2023.

\bibitem[Bach et~al.(2015)Bach, Binder, Montavon, Klauschen, M{\"u}ller, and Samek]{bach2015pixel}
Bach, S., Binder, A., Montavon, G., Klauschen, F., M{\"u}ller, K.-R., and Samek, W.
\newblock On pixel-wise explanations for non-linear classifier decisions by layer-wise relevance propagation.
\newblock \emph{PloS one}, 10\penalty0 (7):\penalty0 e0130140, 2015.

\bibitem[Baehrens et~al.(2010)Baehrens, Schroeter, Harmeling, Kawanabe, Hansen, and M{\"u}ller]{baehrens2010explain}
Baehrens, D., Schroeter, T., Harmeling, S., Kawanabe, M., Hansen, K., and M{\"u}ller, K.-R.
\newblock How to explain individual classification decisions.
\newblock \emph{The Journal of Machine Learning Research}, 11:\penalty0 1803--1831, 2010.

\bibitem[Baniecki et~al.(2021)Baniecki, Kretowicz, Pi{\"A}, Wi{\'L}, et~al.]{baniecki2021dalex}
Baniecki, H., Kretowicz, W., Pi{\"A}, P., Wi{\'L}, J., et~al.
\newblock Dalex: responsible machine learning with interactive explainability and fairness in python.
\newblock \emph{Journal of Machine Learning Research}, 22\penalty0 (214):\penalty0 1--7, 2021.

\bibitem[Bau et~al.(2020)Bau, Zhu, Strobelt, Lapedriza, Zhou, and Torralba]{bau2020understanding}
Bau, D., Zhu, J.-Y., Strobelt, H., Lapedriza, A., Zhou, B., and Torralba, A.
\newblock Understanding the role of individual units in a deep neural network.
\newblock \emph{Proceedings of the National Academy of Sciences}, 117\penalty0 (48):\penalty0 30071--30078, 2020.

\bibitem[Bellamy et~al.(2018)Bellamy, Dey, Hind, Hoffman, Houde, Kannan, Lohia, Martino, Mehta, Mojsilovic, et~al.]{bellamy2018ai}
Bellamy, R.~K., Dey, K., Hind, M., Hoffman, S.~C., Houde, S., Kannan, K., Lohia, P., Martino, J., Mehta, S., Mojsilovic, A., et~al.
\newblock Ai fairness 360: An extensible toolkit for detecting, understanding, and mitigating unwanted algorithmic bias.
\newblock \emph{arXiv preprint arXiv:1810.01943}, 2018.

\bibitem[Bender et~al.(2021)Bender, Gebru, McMillan-Major, and Shmitchell]{bender2021dangers}
Bender, E.~M., Gebru, T., McMillan-Major, A., and Shmitchell, S.
\newblock On the dangers of stochastic parrots: Can language models be too big?
\newblock In \emph{Proceedings of the 2021 ACM conference on fairness, accountability, and transparency}, pp.\  610--623, 2021.

\bibitem[Bengio et~al.(2021)Bengio, Lecun, and Hinton]{bengio2021deep}
Bengio, Y., Lecun, Y., and Hinton, G.
\newblock Deep learning for ai.
\newblock \emph{Communications of the ACM}, 64\penalty0 (7):\penalty0 58--65, 2021.

\bibitem[Bien \& Tibshirani(2011)Bien and Tibshirani]{bien2011prototype}
Bien, J. and Tibshirani, R.
\newblock Prototype selection for interpretable classification.
\newblock 2011.

\bibitem[Bishop(2006)]{bishop2006pattern}
Bishop, C.
\newblock Pattern recognition and machine learning.
\newblock \emph{Springer google schola}, 2:\penalty0 531--537, 2006.

\bibitem[Bj{\"o}rklund et~al.(2023)Bj{\"o}rklund, M{\"a}kel{\"a}, and Puolam{\"a}ki]{bjorklund2023slisemap}
Bj{\"o}rklund, A., M{\"a}kel{\"a}, J., and Puolam{\"a}ki, K.
\newblock Slisemap: Supervised dimensionality reduction through local explanations.
\newblock \emph{Machine Learning}, 112\penalty0 (1):\penalty0 1--43, 2023.

\bibitem[Bostrom(2014)]{bostrom2014superintelligence}
Bostrom, N.
\newblock \emph{Superintelligence: Paths, dangers, strategies}.
\newblock OUP Oxford, 2014.

\bibitem[Breiman(2001)]{breiman2001random}
Breiman, L.
\newblock Random forests.
\newblock \emph{Machine learning}, 45:\penalty0 5--32, 2001.

\bibitem[Brown et~al.(2020)Brown, Mann, Ryder, Subbiah, Kaplan, Dhariwal, Neelakantan, Shyam, Sastry, Askell, et~al.]{brown2020language}
Brown, T., Mann, B., Ryder, N., Subbiah, M., Kaplan, J.~D., Dhariwal, P., Neelakantan, A., Shyam, P., Sastry, G., Askell, A., et~al.
\newblock Language models are few-shot learners.
\newblock \emph{Advances in neural information processing systems}, 33:\penalty0 1877--1901, 2020.

\bibitem[Chamola et~al.(2023)Chamola, Hassija, Sulthana, Ghosh, Dhingra, and Sikdar]{chamola2023review}
Chamola, V., Hassija, V., Sulthana, A.~R., Ghosh, D., Dhingra, D., and Sikdar, B.
\newblock A review of trustworthy and explainable artificial intelligence (xai).
\newblock \emph{IEEE Access}, 2023.

\bibitem[Chattopadhay et~al.(2018)Chattopadhay, Sarkar, Howlader, and Balasubramanian]{chattopadhay2018grad}
Chattopadhay, A., Sarkar, A., Howlader, P., and Balasubramanian, V.~N.
\newblock Grad-cam++: Generalized gradient-based visual explanations for deep convolutional networks.
\newblock In \emph{2018 IEEE winter conference on applications of computer vision (WACV)}, pp.\  839--847. IEEE, 2018.

\bibitem[Clark et~al.(2019)Clark, Pantidi, Cooney, Doyle, Garaialde, Edwards, Spillane, Gilmartin, Murad, Munteanu, et~al.]{clark2019makes}
Clark, L., Pantidi, N., Cooney, O., Doyle, P., Garaialde, D., Edwards, J., Spillane, B., Gilmartin, E., Murad, C., Munteanu, C., et~al.
\newblock What makes a good conversation? challenges in designing truly conversational agents.
\newblock In \emph{Proceedings of the 2019 CHI conference on human factors in computing systems}, pp.\  1--12, 2019.

\bibitem[Dean(2020)]{dean20201}
Dean, J.
\newblock 1.1 the deep learning revolution and its implications for computer architecture and chip design.
\newblock In \emph{2020 IEEE International Solid-State Circuits Conference-(ISSCC)}, pp.\  8--14. IEEE, 2020.

\bibitem[Dehaene(2014)]{dehaene2014consciousness}
Dehaene, S.
\newblock \emph{Consciousness and the brain: Deciphering how the brain codes our thoughts}.
\newblock Penguin, 2014.

\bibitem[Dhariwal \& Nichol(2021)Dhariwal and Nichol]{dhariwal2021diffusion}
Dhariwal, P. and Nichol, A.
\newblock Diffusion models beat gans on image synthesis.
\newblock \emph{Advances in neural information processing systems}, 34:\penalty0 8780--8794, 2021.

\bibitem[Dieber \& Kirrane(2020)Dieber and Kirrane]{dieber2020model}
Dieber, J. and Kirrane, S.
\newblock Why model why? assessing the strengths and limitations of lime.
\newblock \emph{arXiv preprint arXiv:2012.00093}, 2020.

\bibitem[Dignum(2019)]{dignum2019responsible}
Dignum, V.
\newblock \emph{Responsible artificial intelligence: how to develop and use AI in a responsible way}, volume 2156.
\newblock Springer, 2019.

\bibitem[Donoghue(2002)]{donoghue2002connecting}
Donoghue, J.~P.
\newblock Connecting cortex to machines: recent advances in brain interfaces.
\newblock \emph{Nature neuroscience}, 5\penalty0 (Suppl 11):\penalty0 1085--1088, 2002.

\bibitem[Doshi-Velez \& Kim(2017)Doshi-Velez and Kim]{doshi2017towards}
Doshi-Velez, F. and Kim, B.
\newblock Towards a rigorous science of interpretable machine learning.
\newblock \emph{arXiv preprint arXiv:1702.08608}, 2017.

\bibitem[Ehsan et~al.(2021)Ehsan, Liao, Muller, Riedl, and Weisz]{ehsan2021expanding}
Ehsan, U., Liao, Q.~V., Muller, M., Riedl, M.~O., and Weisz, J.~D.
\newblock Expanding explainability: Towards social transparency in ai systems.
\newblock In \emph{Proceedings of the 2021 CHI Conference on Human Factors in Computing Systems}, pp.\  1--19, 2021.

\bibitem[Fetz(2007)]{fetz2007volitional}
Fetz, E.~E.
\newblock Volitional control of neural activity: implications for brain--computer interfaces.
\newblock \emph{The Journal of physiology}, 579\penalty0 (3):\penalty0 571--579, 2007.

\bibitem[Fiorucci et~al.(2020)Fiorucci, Khoroshiltseva, Pontil, Traviglia, Del~Bue, and James]{fiorucci2020machine}
Fiorucci, M., Khoroshiltseva, M., Pontil, M., Traviglia, A., Del~Bue, A., and James, S.
\newblock Machine learning for cultural heritage: A survey.
\newblock \emph{Pattern Recognition Letters}, 133:\penalty0 102--108, 2020.

\bibitem[Friston(2005)]{friston2005models}
Friston, K.~J.
\newblock Models of brain function in neuroimaging.
\newblock \emph{Annu. Rev. Psychol.}, 56:\penalty0 57--87, 2005.

\bibitem[Gamez(2021)]{gamez2021measuring}
Gamez, D.
\newblock Measuring intelligence in natural and artificial systems.
\newblock \emph{Journal of Artificial Intelligence and Consciousness}, 8\penalty0 (02):\penalty0 285--302, 2021.

\bibitem[Garcia-Magarino et~al.(2019)Garcia-Magarino, Muttukrishnan, and Lloret]{garcia2019human}
Garcia-Magarino, I., Muttukrishnan, R., and Lloret, J.
\newblock Human-centric ai for trustworthy iot systems with explainable multilayer perceptrons.
\newblock \emph{IEEE Access}, 7:\penalty0 125562--125574, 2019.

\bibitem[Gerrish(2018)]{gerrish2018smart}
Gerrish, S.
\newblock \emph{How smart machines think}.
\newblock MIT Press, 2018.

\bibitem[Gherman et~al.(2023)Gherman, Abdallah, Pang, Gorochowski, Grierson, and Marucci]{gherman2023bridging}
Gherman, I.~M., Abdallah, Z.~S., Pang, W., Gorochowski, T.~E., Grierson, C.~S., and Marucci, L.
\newblock Bridging the gap between mechanistic biological models and machine learning surrogates.
\newblock \emph{PLoS Computational Biology}, 19\penalty0 (4):\penalty0 e1010988, 2023.

\bibitem[Ghorbani et~al.(2019)Ghorbani, Wexler, Zou, and Kim]{ghorbani2019towards}
Ghorbani, A., Wexler, J., Zou, J.~Y., and Kim, B.
\newblock Towards automatic concept-based explanations.
\newblock \emph{Advances in neural information processing systems}, 32, 2019.

\bibitem[Gilpin et~al.(2018)Gilpin, Bau, Yuan, Bajwa, Specter, and Kagal]{gilpin2018explaining}
Gilpin, L.~H., Bau, D., Yuan, B.~Z., Bajwa, A., Specter, M., and Kagal, L.
\newblock Explaining explanations: An overview of interpretability of machine learning.
\newblock In \emph{2018 IEEE 5th International Conference on data science and advanced analytics (DSAA)}, pp.\  80--89. IEEE, 2018.

\bibitem[Goleman(2020)]{goleman2020emotional}
Goleman, D.
\newblock \emph{Emotional intelligence}.
\newblock Bloomsbury Publishing, 2020.

\bibitem[Graziano(2017)]{graziano2017attention}
Graziano, M.~S.
\newblock The attention schema theory: A foundation for engineering artificial consciousness.
\newblock \emph{Frontiers in Robotics and AI}, 4:\penalty0 60, 2017.

\bibitem[Gunning et~al.(2019)Gunning, Stefik, Choi, Miller, Stumpf, and Yang]{gunning2019xai}
Gunning, D., Stefik, M., Choi, J., Miller, T., Stumpf, S., and Yang, G.-Z.
\newblock Xai—explainable artificial intelligence.
\newblock \emph{Science robotics}, 4\penalty0 (37):\penalty0 eaay7120, 2019.

\bibitem[Hassabis et~al.(2017)Hassabis, Kumaran, Summerfield, and Botvinick]{hassabis2017neuroscience}
Hassabis, D., Kumaran, D., Summerfield, C., and Botvinick, M.
\newblock Neuroscience-inspired artificial intelligence.
\newblock \emph{Neuron}, 95\penalty0 (2):\penalty0 245--258, 2017.

\bibitem[He et~al.(2016)He, Zhang, Ren, and Sun]{he2016deep}
He, K., Zhang, X., Ren, S., and Sun, J.
\newblock Deep residual learning for image recognition.
\newblock In \emph{Proceedings of the IEEE conference on computer vision and pattern recognition}, pp.\  770--778, 2016.

\bibitem[Hearst et~al.(1998)Hearst, Dumais, Osuna, Platt, and Scholkopf]{hearst1998support}
Hearst, M.~A., Dumais, S.~T., Osuna, E., Platt, J., and Scholkopf, B.
\newblock Support vector machines.
\newblock \emph{IEEE Intelligent Systems and their applications}, 13\penalty0 (4):\penalty0 18--28, 1998.

\bibitem[Hedstr{\"o}m et~al.(2023)Hedstr{\"o}m, Weber, Krakowczyk, Bareeva, Motzkus, Samek, Lapuschkin, and H{\"o}hne]{hedstrom2023quantus}
Hedstr{\"o}m, A., Weber, L., Krakowczyk, D., Bareeva, D., Motzkus, F., Samek, W., Lapuschkin, S., and H{\"o}hne, M. M.-C.
\newblock Quantus: An explainable ai toolkit for responsible evaluation of neural network explanations and beyond.
\newblock \emph{Journal of Machine Learning Research}, 24\penalty0 (34):\penalty0 1--11, 2023.

\bibitem[Hindennach et~al.(2023)Hindennach, Shi, Mileti{\'c}, and Bulling]{hindennach2023mindful}
Hindennach, S., Shi, L., Mileti{\'c}, F., and Bulling, A.
\newblock Mindful explanations: Prevalence and impact of mind attribution in xai research.
\newblock \emph{arXiv preprint arXiv:2312.12119}, 2023.

\bibitem[Hinton et~al.(2006)Hinton, Osindero, and Teh]{hinton2006fast}
Hinton, G.~E., Osindero, S., and Teh, Y.-W.
\newblock A fast learning algorithm for deep belief nets.
\newblock \emph{Neural computation}, 18\penalty0 (7):\penalty0 1527--1554, 2006.

\bibitem[Ho et~al.(2020)Ho, Jain, and Abbeel]{ho2020denoising}
Ho, J., Jain, A., and Abbeel, P.
\newblock Denoising diffusion probabilistic models.
\newblock \emph{Advances in neural information processing systems}, 33:\penalty0 6840--6851, 2020.

\bibitem[Holzinger et~al.(2017)Holzinger, Biemann, Pattichis, and Kell]{holzinger2017we}
Holzinger, A., Biemann, C., Pattichis, C.~S., and Kell, D.~B.
\newblock What do we need to build explainable ai systems for the medical domain?
\newblock \emph{arXiv preprint arXiv:1712.09923}, 2017.

\bibitem[Hyder et~al.(2019)Hyder, Siau, and Nah]{hyder2019artificial}
Hyder, Z., Siau, K., and Nah, F.
\newblock Artificial intelligence, machine learning, and autonomous technologies in mining industry.
\newblock \emph{Journal of Database Management (JDM)}, 30\penalty0 (2):\penalty0 67--79, 2019.

\bibitem[Ibrahim et~al.(2019)Ibrahim, Louie, Modarres, and Paisley]{ibrahim2019global}
Ibrahim, M., Louie, M., Modarres, C., and Paisley, J.
\newblock Global explanations of neural networks: Mapping the landscape of predictions.
\newblock In \emph{Proceedings of the 2019 AAAI/ACM Conference on AI, Ethics, and Society}, pp.\  279--287, 2019.

\bibitem[Jacovi \& Goldberg(2020)Jacovi and Goldberg]{jacovi2020towards}
Jacovi, A. and Goldberg, Y.
\newblock Towards faithfully interpretable nlp systems: How should we define and evaluate faithfulness?
\newblock \emph{arXiv preprint arXiv:2004.03685}, 2020.

\bibitem[Jayaraman \& Evans(2019)Jayaraman and Evans]{jayaraman2019evaluating}
Jayaraman, B. and Evans, D.
\newblock Evaluating differentially private machine learning in practice.
\newblock In \emph{28th USENIX Security Symposium (USENIX Security 19)}, pp.\  1895--1912, 2019.

\bibitem[Kabir et~al.(2022)Kabir, Abdar, Khosravi, Jalali, Atiya, Nahavandi, and Srinivasan]{kabir2022spinalnet}
Kabir, H.~D., Abdar, M., Khosravi, A., Jalali, S. M.~J., Atiya, A.~F., Nahavandi, S., and Srinivasan, D.
\newblock Spinalnet: Deep neural network with gradual input.
\newblock \emph{IEEE Transactions on Artificial Intelligence}, 2022.

\bibitem[Kim et~al.(2018)Kim, Wattenberg, Gilmer, Cai, Wexler, Viegas, et~al.]{kim2018interpretability}
Kim, B., Wattenberg, M., Gilmer, J., Cai, C., Wexler, J., Viegas, F., et~al.
\newblock Interpretability beyond feature attribution: Quantitative testing with concept activation vectors (tcav).
\newblock In \emph{International conference on machine learning}, pp.\  2668--2677. PMLR, 2018.

\bibitem[Kriegeskorte(2015)]{kriegeskorte2015deep}
Kriegeskorte, N.
\newblock Deep neural networks: a new framework for modeling biological vision and brain information processing.
\newblock \emph{Annual review of vision science}, 1:\penalty0 417--446, 2015.

\bibitem[Kuhl et~al.(2023)Kuhl, Artelt, and Hammer]{kuhl2023let}
Kuhl, U., Artelt, A., and Hammer, B.
\newblock Let's go to the alien zoo: Introducing an experimental framework to study usability of counterfactual explanations for machine learning.
\newblock \emph{Frontiers in Computer Science}, 5:\penalty0 20, 2023.

\bibitem[Lake et~al.(2017)Lake, Ullman, Tenenbaum, and Gershman]{lake2017building}
Lake, B.~M., Ullman, T.~D., Tenenbaum, J.~B., and Gershman, S.~J.
\newblock Building machines that learn and think like people.
\newblock \emph{Behavioral and brain sciences}, 40:\penalty0 e253, 2017.

\bibitem[LeCun et~al.(2015)LeCun, Bengio, and Hinton]{lecun2015deep}
LeCun, Y., Bengio, Y., and Hinton, G.
\newblock Deep learning.
\newblock \emph{nature}, 521\penalty0 (7553):\penalty0 436--444, 2015.

\bibitem[Li et~al.(2019)Li, Khodak, Caldas, and Talwalkar]{li2019differentially}
Li, J., Khodak, M., Caldas, S., and Talwalkar, A.
\newblock Differentially private meta-learning.
\newblock \emph{arXiv preprint arXiv:1909.05830}, 2019.

\bibitem[Linzen(2020)]{linzen2020can}
Linzen, T.
\newblock How can we accelerate progress towards human-like linguistic generalization?
\newblock \emph{arXiv preprint arXiv:2005.00955}, 2020.

\bibitem[Lundberg \& Lee(2017)Lundberg and Lee]{lundberg2017unified}
Lundberg, S.~M. and Lee, S.-I.
\newblock A unified approach to interpreting model predictions.
\newblock \emph{Advances in neural information processing systems}, 30, 2017.

\bibitem[Luxton(2014)]{luxton2014artificial}
Luxton, D.~D.
\newblock Artificial intelligence in psychological practice: Current and future applications and implications.
\newblock \emph{Professional Psychology: Research and Practice}, 45\penalty0 (5):\penalty0 332, 2014.

\bibitem[Marblestone et~al.(2016)Marblestone, Wayne, and Kording]{marblestone2016toward}
Marblestone, A.~H., Wayne, G., and Kording, K.~P.
\newblock Toward an integration of deep learning and neuroscience.
\newblock \emph{Frontiers in computational neuroscience}, 10:\penalty0 94, 2016.

\bibitem[Marcus(2018)]{marcus2018deep}
Marcus, G.
\newblock Deep learning: A critical appraisal.
\newblock \emph{arXiv preprint arXiv:1801.00631}, 2018.

\bibitem[Martin-Brualla et~al.(2021)Martin-Brualla, Radwan, Sajjadi, Barron, Dosovitskiy, and Duckworth]{martin2021nerf}
Martin-Brualla, R., Radwan, N., Sajjadi, M.~S., Barron, J.~T., Dosovitskiy, A., and Duckworth, D.
\newblock Nerf in the wild: Neural radiance fields for unconstrained photo collections.
\newblock In \emph{Proceedings of the IEEE/CVF Conference on Computer Vision and Pattern Recognition}, pp.\  7210--7219, 2021.

\bibitem[McDuff \& Czerwinski(2018)McDuff and Czerwinski]{mcduff2018designing}
McDuff, D. and Czerwinski, M.
\newblock Designing emotionally sentient agents.
\newblock \emph{Communications of the ACM}, 61\penalty0 (12):\penalty0 74--83, 2018.

\bibitem[McLean et~al.(2023)McLean, Read, Thompson, Baber, Stanton, and Salmon]{mclean2023risks}
McLean, S., Read, G.~J., Thompson, J., Baber, C., Stanton, N.~A., and Salmon, P.~M.
\newblock The risks associated with artificial general intelligence: A systematic review.
\newblock \emph{Journal of Experimental \& Theoretical Artificial Intelligence}, 35\penalty0 (5):\penalty0 649--663, 2023.

\bibitem[McStay(2018)]{mcstay2018emotional}
McStay, A.
\newblock Emotional ai: The rise of empathic media.
\newblock \emph{Emotional AI}, pp.\  1--248, 2018.

\bibitem[Mehrabi et~al.(2021)Mehrabi, Morstatter, Saxena, Lerman, and Galstyan]{mehrabi2021survey}
Mehrabi, N., Morstatter, F., Saxena, N., Lerman, K., and Galstyan, A.
\newblock A survey on bias and fairness in machine learning.
\newblock \emph{ACM computing surveys (CSUR)}, 54\penalty0 (6):\penalty0 1--35, 2021.

\bibitem[Mildenhall et~al.(2021)Mildenhall, Srinivasan, Tancik, Barron, Ramamoorthi, and Ng]{mildenhall2021nerf}
Mildenhall, B., Srinivasan, P.~P., Tancik, M., Barron, J.~T., Ramamoorthi, R., and Ng, R.
\newblock Nerf: Representing scenes as neural radiance fields for view synthesis.
\newblock \emph{Communications of the ACM}, 65\penalty0 (1):\penalty0 99--106, 2021.

\bibitem[Miller(2019)]{miller2019explanation}
Miller, T.
\newblock Explanation in artificial intelligence: Insights from the social sciences.
\newblock \emph{Artificial intelligence}, 267:\penalty0 1--38, 2019.

\bibitem[Mittelstadt(2019)]{mittelstadt2019principles}
Mittelstadt, B.
\newblock Principles alone cannot guarantee ethical ai.
\newblock \emph{Nature machine intelligence}, 1\penalty0 (11):\penalty0 501--507, 2019.

\bibitem[Moura \& Carro(2023)Moura and Carro]{moura2023scalable}
Moura, R. F.~d. and Carro, L.
\newblock Scalable and energy-efficient nn acceleration with gpu-reram architecture.
\newblock In \emph{International Symposium on Applied Reconfigurable Computing}, pp.\  230--244. Springer, 2023.

\bibitem[Moxon \& Foffani(2015)Moxon and Foffani]{moxon2015brain}
Moxon, K.~A. and Foffani, G.
\newblock Brain-machine interfaces beyond neuroprosthetics.
\newblock \emph{Neuron}, 86\penalty0 (1):\penalty0 55--67, 2015.

\bibitem[Noh et~al.(2015)Noh, Hong, and Han]{noh2015learning}
Noh, H., Hong, S., and Han, B.
\newblock Learning deconvolution network for semantic segmentation.
\newblock In \emph{Proceedings of the IEEE international conference on computer vision}, pp.\  1520--1528, 2015.

\bibitem[Peterson(2009)]{peterson2009k}
Peterson, L.~E.
\newblock K-nearest neighbor.
\newblock \emph{Scholarpedia}, 4\penalty0 (2):\penalty0 1883, 2009.

\bibitem[Pope et~al.(2019)Pope, Kolouri, Rostami, Martin, and Hoffmann]{pope2019explainability}
Pope, P.~E., Kolouri, S., Rostami, M., Martin, C.~E., and Hoffmann, H.
\newblock Explainability methods for graph convolutional neural networks.
\newblock In \emph{Proceedings of the IEEE/CVF conference on computer vision and pattern recognition}, pp.\  10772--10781, 2019.

\bibitem[Premack \& Woodruff(1978)Premack and Woodruff]{premack1978does}
Premack, D. and Woodruff, G.
\newblock Does the chimpanzee have a theory of mind?
\newblock \emph{Behavioral and brain sciences}, 1\penalty0 (4):\penalty0 515--526, 1978.

\bibitem[Prieto et~al.(2016)Prieto, Prieto, Ortigosa, Ros, Pelayo, Ortega, and Rojas]{prieto2016neural}
Prieto, A., Prieto, B., Ortigosa, E.~M., Ros, E., Pelayo, F., Ortega, J., and Rojas, I.
\newblock Neural networks: An overview of early research, current frameworks and new challenges.
\newblock \emph{Neurocomputing}, 214:\penalty0 242--268, 2016.

\bibitem[Quinlan et~al.(1996)]{quinlan1996bagging}
Quinlan, J.~R. et~al.
\newblock Bagging, boosting, and c4. 5.
\newblock In \emph{Aaai/Iaai, vol. 1}, pp.\  725--730. Citeseer, 1996.

\bibitem[Rahwan et~al.(2019)Rahwan, Cebrian, Obradovich, Bongard, Bonnefon, Breazeal, Crandall, Christakis, Couzin, Jackson, et~al.]{rahwan2019machine}
Rahwan, I., Cebrian, M., Obradovich, N., Bongard, J., Bonnefon, J.-F., Breazeal, C., Crandall, J.~W., Christakis, N.~A., Couzin, I.~D., Jackson, M.~O., et~al.
\newblock Machine behaviour.
\newblock \emph{Nature}, 568\penalty0 (7753):\penalty0 477--486, 2019.

\bibitem[Ribeiro et~al.(2016)Ribeiro, Singh, and Guestrin]{ribeiro2016should}
Ribeiro, M.~T., Singh, S., and Guestrin, C.
\newblock " why should i trust you?" explaining the predictions of any classifier.
\newblock In \emph{Proceedings of the 22nd ACM SIGKDD international conference on knowledge discovery and data mining}, pp.\  1135--1144, 2016.

\bibitem[Riek(2017)]{riek2017healthcare}
Riek, L.~D.
\newblock Healthcare robotics.
\newblock \emph{Communications of the ACM}, 60\penalty0 (11):\penalty0 68--78, 2017.

\bibitem[Rish et~al.(2001)]{rish2001empirical}
Rish, I. et~al.
\newblock An empirical study of the naive bayes classifier.
\newblock In \emph{IJCAI 2001 workshop on empirical methods in artificial intelligence}, volume~3, pp.\  41--46, 2001.

\bibitem[Ritchie et~al.(2019)Ritchie, Kaplan, and Klein]{ritchie2019decoding}
Ritchie, J.~B., Kaplan, D.~M., and Klein, C.
\newblock Decoding the brain: Neural representation and the limits of multivariate pattern analysis in cognitive neuroscience.
\newblock \emph{The British journal for the philosophy of science}, 2019.

\bibitem[Russell(2019)]{russell2019human}
Russell, S.
\newblock \emph{Human compatible: Artificial intelligence and the problem of control}.
\newblock Penguin, 2019.

\bibitem[Saeed \& Omlin(2023)Saeed and Omlin]{saeed2023explainable}
Saeed, W. and Omlin, C.
\newblock Explainable ai (xai): A systematic meta-survey of current challenges and future opportunities.
\newblock \emph{Knowledge-Based Systems}, 263:\penalty0 110273, 2023.

\bibitem[Saleem et~al.(2022)Saleem, Senan, Wahid, Aamir, Samad, Khan, et~al.]{saleem2022comparative}
Saleem, M.~A., Senan, N., Wahid, F., Aamir, M., Samad, A., Khan, M., et~al.
\newblock Comparative analysis of recent architecture of convolutional neural network.
\newblock \emph{Mathematical Problems in Engineering}, 2022, 2022.

\bibitem[Samek et~al.(2019)Samek, Montavon, Vedaldi, Hansen, and M{\"u}ller]{samek2019explainable}
Samek, W., Montavon, G., Vedaldi, A., Hansen, L.~K., and M{\"u}ller, K.-R.
\newblock \emph{Explainable AI: interpreting, explaining and visualizing deep learning}, volume 11700.
\newblock Springer Nature, 2019.

\bibitem[Schmid \& Wrede(2022)Schmid and Wrede]{schmid2022missing}
Schmid, U. and Wrede, B.
\newblock What is missing in xai so far? an interdisciplinary perspective.
\newblock \emph{KI-K{\"u}nstliche Intelligenz}, 36\penalty0 (3-4):\penalty0 303--315, 2022.

\bibitem[Segal \& Demos(2018)Segal and Demos]{segal2018gender}
Segal, M.~T. and Demos, V.
\newblock \emph{Gender and the media: Women’s places}.
\newblock Emerald Publishing Limited, 2018.

\bibitem[Selvaraju et~al.(2017)Selvaraju, Cogswell, Das, Vedantam, Parikh, and Batra]{selvaraju2017grad}
Selvaraju, R.~R., Cogswell, M., Das, A., Vedantam, R., Parikh, D., and Batra, D.
\newblock Grad-cam: Visual explanations from deep networks via gradient-based localization.
\newblock In \emph{Proceedings of the IEEE international conference on computer vision}, pp.\  618--626, 2017.

\bibitem[Shneiderman(2022)]{shneiderman2022human}
Shneiderman, B.
\newblock \emph{Human-centered AI}.
\newblock Oxford University Press, 2022.

\bibitem[Shrikumar et~al.(2017)Shrikumar, Greenside, and Kundaje]{shrikumar2017learning}
Shrikumar, A., Greenside, P., and Kundaje, A.
\newblock Learning important features through propagating activation differences.
\newblock In \emph{International conference on machine learning}, pp.\  3145--3153. PMLR, 2017.

\bibitem[Simonyan \& Zisserman(2014)Simonyan and Zisserman]{simonyan2014very}
Simonyan, K. and Zisserman, A.
\newblock Very deep convolutional networks for large-scale image recognition.
\newblock \emph{arXiv preprint arXiv:1409.1556}, 2014.

\bibitem[Simonyan et~al.(2013)Simonyan, Vedaldi, and Zisserman]{simonyan2013deep}
Simonyan, K., Vedaldi, A., and Zisserman, A.
\newblock Deep inside convolutional networks: Visualising image classification models and saliency maps.
\newblock \emph{arXiv preprint arXiv:1312.6034}, 2013.

\bibitem[Slack et~al.(2021)Slack, Hilgard, Singh, and Lakkaraju]{slack2021reliable}
Slack, D., Hilgard, A., Singh, S., and Lakkaraju, H.
\newblock Reliable post hoc explanations: Modeling uncertainty in explainability.
\newblock \emph{Advances in neural information processing systems}, 34:\penalty0 9391--9404, 2021.

\bibitem[Suffian et~al.(2023)Suffian, Kuhl, Alonso-Moral, and Bogliolo]{suffian2023toward}
Suffian, M., Kuhl, U., Alonso-Moral, J.~M., and Bogliolo, A.
\newblock Toward enriched cognitive learning with xai.
\newblock \emph{arXiv preprint arXiv:2312.12290}, 2023.

\bibitem[Sundararajan et~al.(2017)Sundararajan, Taly, and Yan]{sundararajan2017axiomatic}
Sundararajan, M., Taly, A., and Yan, Q.
\newblock Axiomatic attribution for deep networks.
\newblock In \emph{International conference on machine learning}, pp.\  3319--3328. PMLR, 2017.

\bibitem[Tewari et~al.(2022)Tewari, Thies, Mildenhall, Srinivasan, Tretschk, Yifan, Lassner, Sitzmann, Martin-Brualla, Lombardi, et~al.]{tewari2022advances}
Tewari, A., Thies, J., Mildenhall, B., Srinivasan, P., Tretschk, E., Yifan, W., Lassner, C., Sitzmann, V., Martin-Brualla, R., Lombardi, S., et~al.
\newblock Advances in neural rendering.
\newblock In \emph{Computer Graphics Forum}, volume~41, pp.\  703--735. Wiley Online Library, 2022.

\bibitem[Tononi et~al.(2016)Tononi, Boly, Massimini, and Koch]{tononi2016integrated}
Tononi, G., Boly, M., Massimini, M., and Koch, C.
\newblock Integrated information theory: from consciousness to its physical substrate.
\newblock \emph{Nature Reviews Neuroscience}, 17\penalty0 (7):\penalty0 450--461, 2016.

\bibitem[Tsoukalas(2018)]{tsoukalas2018theory}
Tsoukalas, I.
\newblock Theory of mind: towards an evolutionary theory.
\newblock \emph{Evolutionary Psychological Science}, 4\penalty0 (1):\penalty0 38--66, 2018.

\bibitem[Wachter et~al.(2017)Wachter, Mittelstadt, and Russell]{wachter2017counterfactual}
Wachter, S., Mittelstadt, B., and Russell, C.
\newblock Counterfactual explanations without opening the black box: Automated decisions and the gdpr.
\newblock \emph{Harv. JL \& Tech.}, 31:\penalty0 841, 2017.

\bibitem[Wallach \& Allen(2008)Wallach and Allen]{wallach2008moral}
Wallach, W. and Allen, C.
\newblock \emph{Moral machines: Teaching robots right from wrong}.
\newblock Oxford University Press, 2008.

\bibitem[Wang et~al.(2019)Wang, She, and Ward]{wang2019generative}
Wang, Z., She, Q., and Ward, T.~E.
\newblock Generative adversarial networks: A survey and taxonomy.
\newblock \emph{arXiv preprint arXiv:1906.01529}, 2, 2019.

\bibitem[Wang et~al.(2020)Wang, She, Smeaton, Ward, and Healy]{wang2020synthetic}
Wang, Z., She, Q., Smeaton, A.~F., Ward, T.~E., and Healy, G.
\newblock Synthetic-neuroscore: Using a neuro-ai interface for evaluating generative adversarial networks.
\newblock \emph{Neurocomputing}, 405:\penalty0 26--36, 2020.

\bibitem[Wang et~al.(2022)Wang, Liu, Luo, Xu, Xie, Wang, Cai, Qi, Yuan, Yang, et~al.]{wang2022advanced}
Wang, Z., Liu, M., Luo, Y., Xu, Z., Xie, Y., Wang, L., Cai, L., Qi, Q., Yuan, Z., Yang, T., et~al.
\newblock Advanced graph and sequence neural networks for molecular property prediction and drug discovery.
\newblock \emph{Bioinformatics}, 38\penalty0 (9):\penalty0 2579--2586, 2022.

\bibitem[Weerts et~al.(2023)Weerts, Dud{\'\i}k, Edgar, Jalali, Lutz, and Madaio]{weerts2023fairlearn}
Weerts, H., Dud{\'\i}k, M., Edgar, R., Jalali, A., Lutz, R., and Madaio, M.
\newblock Fairlearn: Assessing and improving fairness of ai systems.
\newblock \emph{arXiv preprint arXiv:2303.16626}, 2023.

\bibitem[Wellman(2014)]{wellman2014making}
Wellman, H.~M.
\newblock \emph{Making minds: How theory of mind develops}.
\newblock Oxford University Press, 2014.

\bibitem[Wiegreffe \& Pinter(2019)Wiegreffe and Pinter]{wiegreffe2019attention}
Wiegreffe, S. and Pinter, Y.
\newblock Attention is not not explanation.
\newblock \emph{arXiv preprint arXiv:1908.04626}, 2019.

\bibitem[Yamins \& DiCarlo(2016)Yamins and DiCarlo]{yamins2016using}
Yamins, D.~L. and DiCarlo, J.~J.
\newblock Using goal-driven deep learning models to understand sensory cortex.
\newblock \emph{Nature neuroscience}, 19\penalty0 (3):\penalty0 356--365, 2016.

\bibitem[Yuan et~al.(2022)Yuan, Yu, Gui, and Ji]{yuan2022explainability}
Yuan, H., Yu, H., Gui, S., and Ji, S.
\newblock Explainability in graph neural networks: A taxonomic survey.
\newblock \emph{IEEE transactions on pattern analysis and machine intelligence}, 45\penalty0 (5):\penalty0 5782--5799, 2022.

\bibitem[Zall \& Kangavari(2022)Zall and Kangavari]{zall2022comparative}
Zall, R. and Kangavari, M.~R.
\newblock Comparative analytical survey on cognitive agents with emotional intelligence.
\newblock \emph{Cognitive Computation}, 14\penalty0 (4):\penalty0 1223--1246, 2022.

\bibitem[Zhan et~al.(2023)Zhan, Yu, Wu, Zhang, Lu, Liu, Kortylewski, Theobalt, and Xing]{zhan2023multimodal}
Zhan, F., Yu, Y., Wu, R., Zhang, J., Lu, S., Liu, L., Kortylewski, A., Theobalt, C., and Xing, E.
\newblock Multimodal image synthesis and editing: A survey and taxonomy.
\newblock \emph{IEEE Transactions on Pattern Analysis and Machine Intelligence}, 2023.

\bibitem[Zhang et~al.(2023)Zhang, Zhu, and Su]{zhang2023toward}
Zhang, B., Zhu, J., and Su, H.
\newblock Toward the third generation artificial intelligence.
\newblock \emph{Science China Information Sciences}, 66\penalty0 (2):\penalty0 121101, 2023.

\bibitem[Zhou et~al.(2016)Zhou, Khosla, Lapedriza, Oliva, and Torralba]{zhou2016learning}
Zhou, B., Khosla, A., Lapedriza, A., Oliva, A., and Torralba, A.
\newblock Learning deep features for discriminative localization.
\newblock In \emph{Proceedings of the IEEE conference on computer vision and pattern recognition}, pp.\  2921--2929, 2016.

\end{thebibliography}
\bibliographystyle{icml2024}




\end{document}